\documentclass[letterpaper, 10 pt, conference]{ieeeconf}  

\IEEEoverridecommandlockouts                              

\usepackage{cite}
\usepackage{amsmath,amssymb,amsfonts}
\usepackage{algorithmic}
\usepackage{graphicx}
\usepackage{textcomp}
\usepackage{xcolor}
\usepackage{float}
\usepackage{subcaption}
\usepackage{multirow}

\title{\LARGE \bf
What the HoloLens Maps Is Your Workspace: Fast Mapping and Set-up of Robot Cells via Head Mounted Displays and Augmented Reality
}

\author{David Puljiz$^{1}$, Franziska Krebs$^{2}$, Fabian B\"osing$^{1}$, Bj\"orn Hein $^{1,3}$
\thanks{$^{1}$Intelligent Process Automation and Robotics Lab (IPR), Institute for Anthropomatics and Robotics, Karlsruhe Institute of Technology, Karlsruhe, Germany
        {\tt\small david.puljiz@kit.edu}}%
\thanks{$^{2}$ High Performance Humanoid Technologies Lab (H²T), Institute for Anthropomatics and Robotics, Karlsruhe Institute of Technology, Karlsruhe, Germany}%
\thanks{$^{3}$ Karlsruhe University of Applied Sciences, Karlsruhe, Germany}%
}

\begin{document}

\maketitle
\thispagestyle{empty}
\pagestyle{empty}

\begin{abstract}

Classical methods of modelling and mapping robot work cells are time consuming, expensive and involve expert knowledge. We present a novel approach to mapping and cell setup using modern Head Mounted Displays (HMDs) that possess self-localisation and mapping capabilities. We leveraged these capabilities to create a point cloud of the environment and build an OctoMap - a voxel occupancy grid representation of the robot's workspace for path planning. Through the use of  Augmented Reality (AR) interactions, the user can edit the created Octomap and add security zones. We perform comprehensive tests of the HoloLens' depth sensing capabilities and the quality of the resultant point cloud. A high-end laser scanner is used to provide the ground truth for the evaluation of the point cloud quality. The amount of false-positive and false-negative voxels in the OctoMap are also tested.  

\end{abstract}
\begin{figure}[ht!]

    \includegraphics[width=0.48\textwidth]{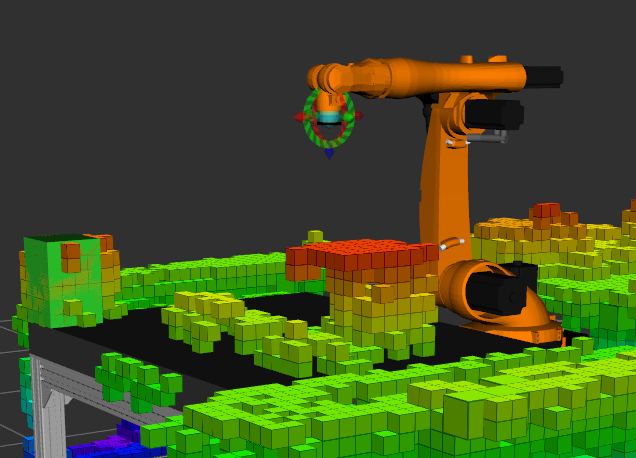}
    \newline
    \includegraphics[width=0.48\textwidth]{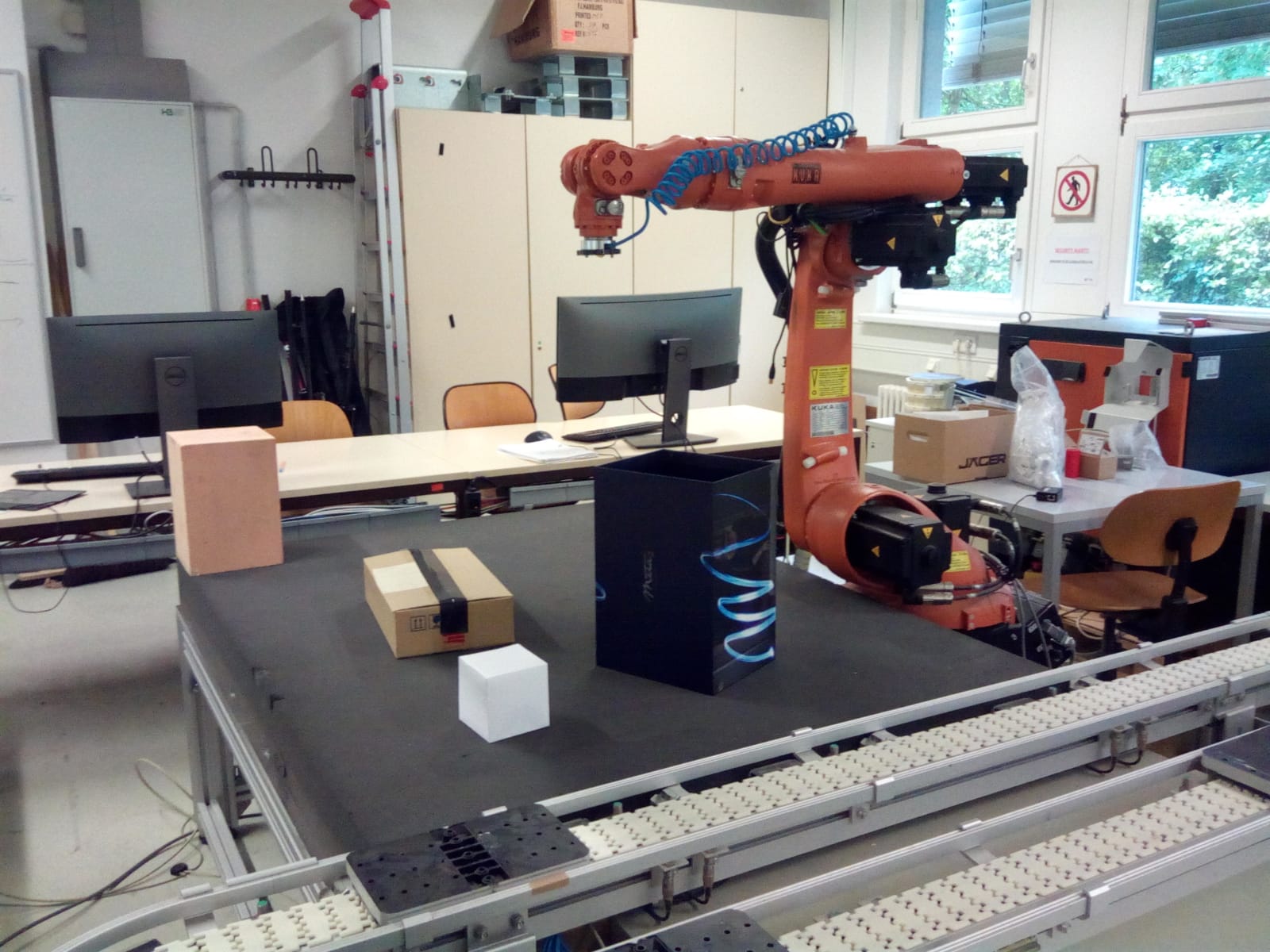}
    \caption[]{Bottom: The real scene mapped with the HoloLens. Top: The resulting OctoMap after the HoloLens mapped the environment and the point cloud was filtered and converted to a voxel occupancy representation.}
    \label{fig:intro}
\end{figure}

\section{INTRODUCTION}
\label{sec:intro}

Knowledge of the robot environment is essential both in offline programming and for auto-generated trajectories as most manipulators lack external sensors to allow them the ability to map their own environment. \par 

Programming robotic manipulators is classically a time consuming process requiring expert knowledge. Offline programming is generally preferred to online, lead-through programming due to smaller downtimes \cite{Pan2012ProgrammingMethods}. Offline programming, however, requires a precise model of the working environment which is often a time-consuming undertaking requiring exact 3D models of the objects around the robot and precise measurements of their placement. Even then, the final program needs to be tested and verified inside the real workspace itself. As it requires significant financial investment, expert knowledge and long delivery times, offline programming is unsuited for small and medium enterprises which require intuitive and fast robot programming paradigms \cite{schraft2006need}. \par   

Likewise setting up safety zones is a time consuming process, mostly done offline and then checked and rechecked until all safety zones are validated. \par

In this paper we propose a cost-efficient method to set-up the working environment of the robot that doesn't require any expert knowledge and combines exceedingly well with newer, AR programming paradigms such as the one presented in \cite{Quintero2018ARProg}. It leverages the localisation and depth sensing capabilities of modern HMDs to map the workspace of the robot. This map is then represented as an OctoMap - a 3D occupancy grid of voxels. The user is then able to add safety zones in situ and edit the OctoMap to, for example, allow collisions in parts where the use case requires contact with specific surfaces. \par 

\begin{figure*}[ht]
\centering
\vspace {5pt}
\subcaptionbox{}
[0.32\textwidth]{ \includegraphics[width=0.32\textwidth]{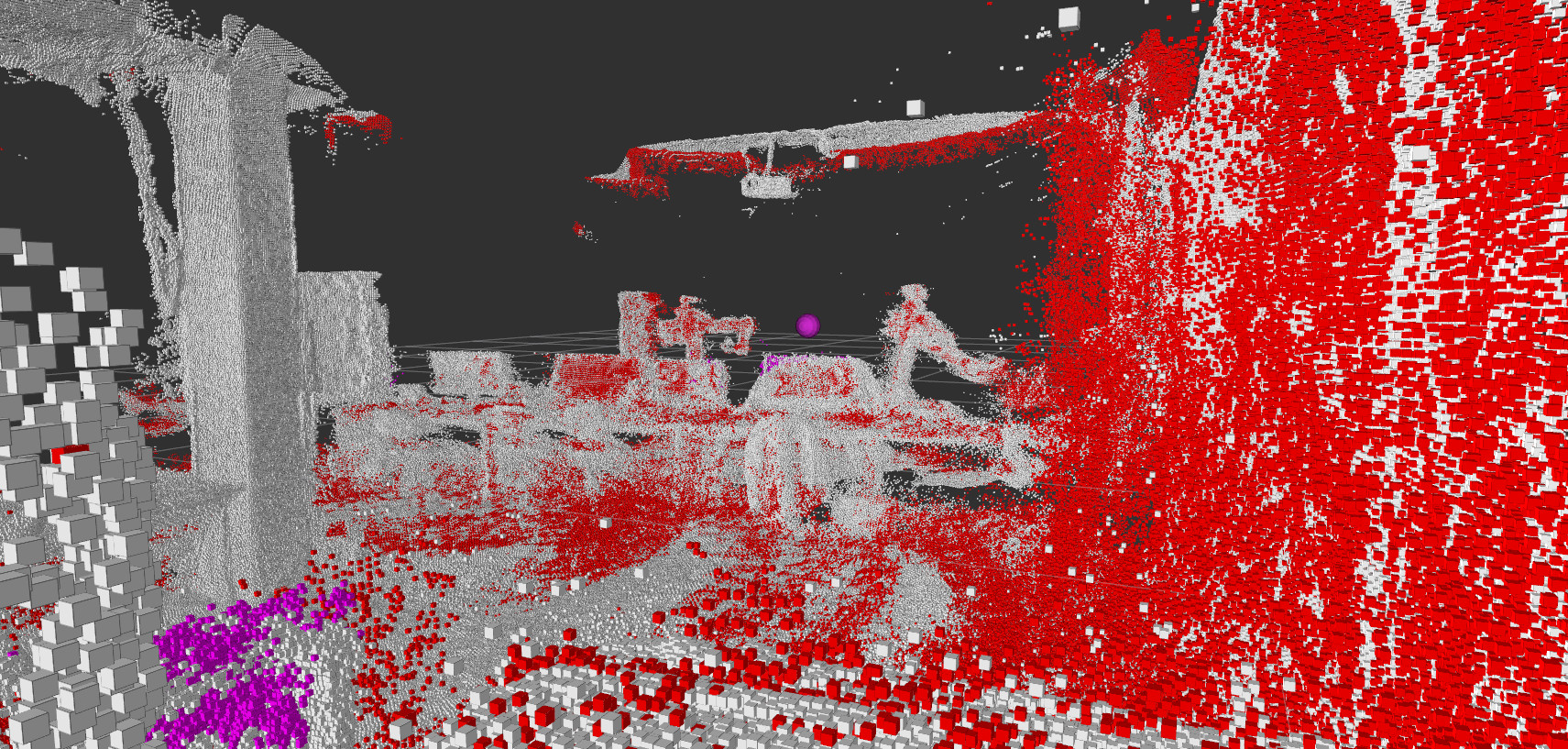}}
\subcaptionbox{}
[0.32\textwidth]{\includegraphics[width=0.32\textwidth]{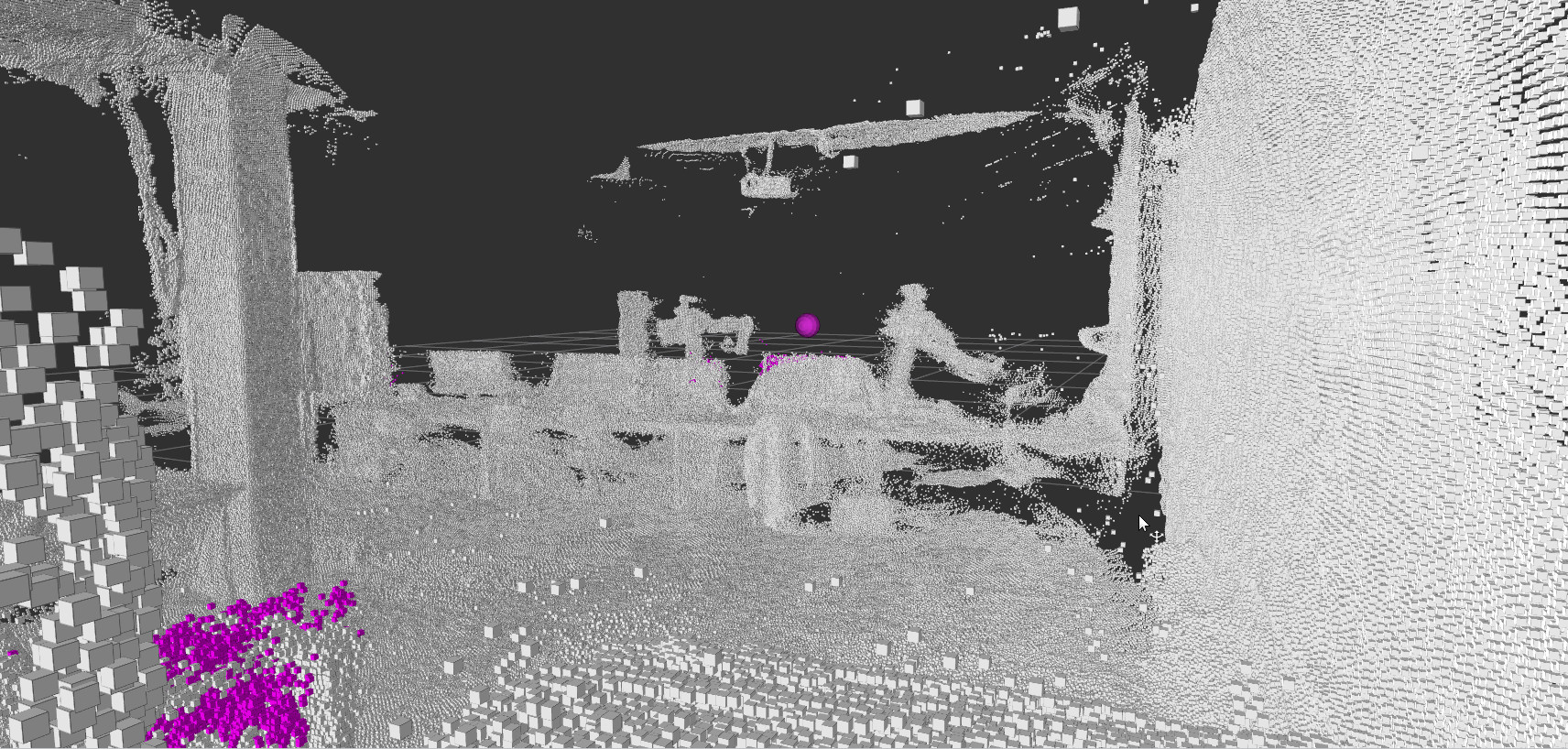}}
\subcaptionbox{}
[0.32\textwidth]{\includegraphics[width=0.32\textwidth]{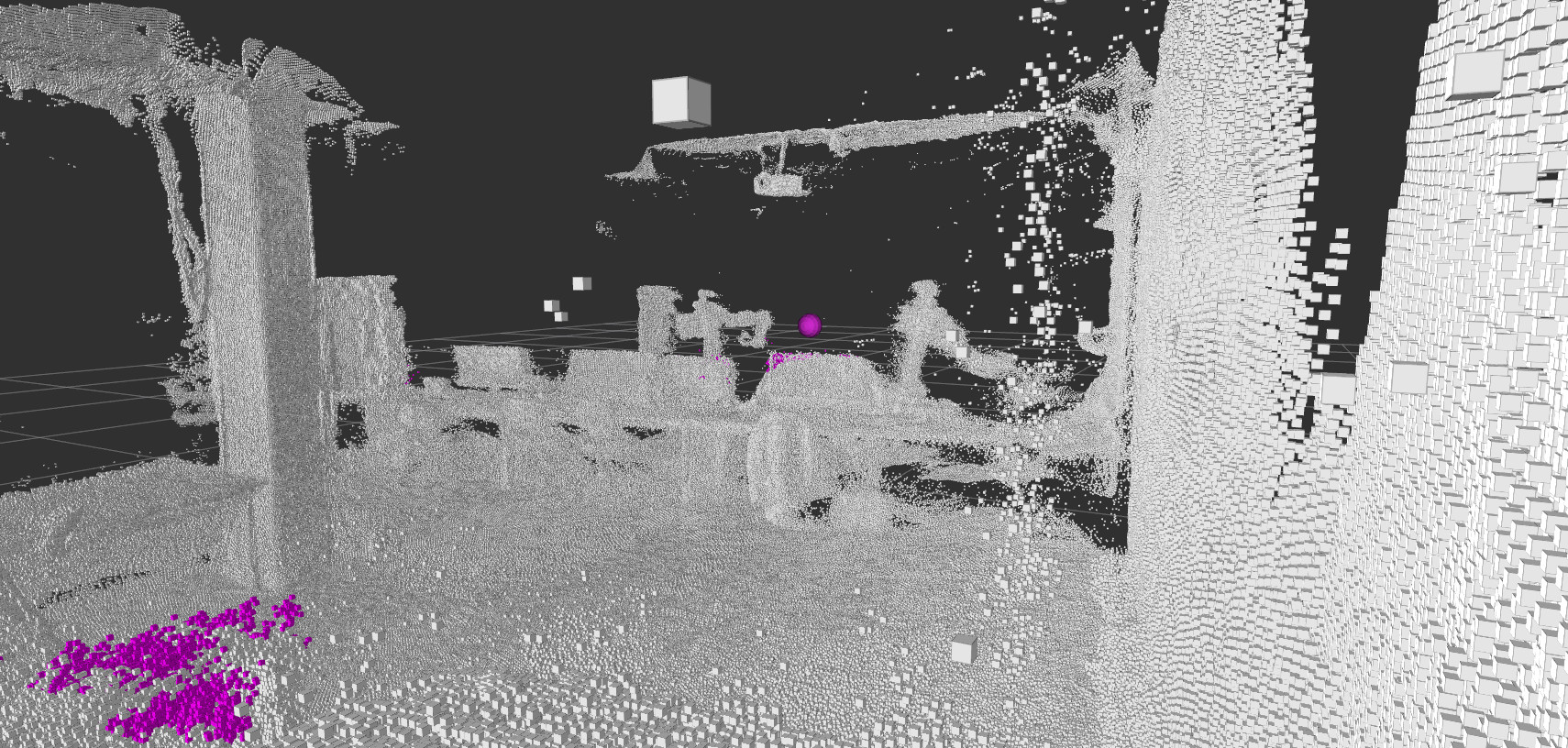}}
    \caption[]{(a) The mapped point cloud with points more than 3.3 meter distant from the depth sensor at the time of capture marked in red. Points at less than 1 meter are marked in purple; (b) Point cloud with points more than 3.3 meter distant removed. One can see the presence of sparse outliers that can be filtered out; (c) Point cloud with points more than 3.4 meters away removed. The outliers are much denser requiring more aggressive filtering which may degrade the quality of the inliers.}
    \label{fig:cutoff}
\end{figure*}

The structure of this paper is as follows. In Section \ref{sec:related work} the state of the art and related work is presented. The contribution of this paper to the state of the art is outlined. Section \ref{sec:methods} describes how the point cloud is constructed (Section \ref{sec:mapping}, how the coordinate transform between the robot and the HoloLens coordinate systems is obtained (Section \ref{sec:referencing}), and finally how the OctoMap is built and edited (Section \ref{sec:octomap}). In Section \ref{sec:experiments} the experiments used to validate our approach are described and the results discussed. The paper ends with Section \ref{sec:conclusion} where the conclusions are given and future improvements are outlined. \par 


\section{RELATED WORK}
\label{sec:related work}

Although most industrial robot manufacturing companies offer software for offline programming of robots, such as ABB's RobotStudio or KUKA's KUKA.Sim, these require precise CAD models of all objects in the environment as well as exact calibration between the virtual and the real robot cell. \par

Neto et al. \cite{NETO2013DirectOffLine} describe a more intuitive offline programming method based on the common CAD package Autodesk Inventor. The user inputs tool coordinates and a program is automatically created. This still requires precise CAD models and calibration. They note that calibration errors are a major source of inaccuracies. According to the authors calibration requires expensive measurement hardware, software and expert knowledge. They also note that external sensing can help mitigate the errors of offline programming. \par

In \cite{Vincze2003Detection} a trajectory is auto-generated for a spray-painting task by using range images of the part to colour. The collision-free trajectory generation, however, still required a model of the robot cell. \par

In the field of AR-based robotics, several approaches exist to plan robot motion in unknown environments. Ong et al. \cite{ong2010novel} use a tracked pointer tool to manually input trajectories and define collision-free volumes. However, no map of the environment is created and only a small part of the total collision-free volume is used. The authors themselves note that alternative methods for generating collision-free volumes should be explored. \par

Similarly, Quintero et al.\cite{Quintero2018ARProg} use holographic waypoints and B-spline interpolation to plan robot trajectories. The system relies on the user to manually modify trajectories to avoid obstacles. The mesh of the environment generated by the HoloLens is used to define waypoints on surfaces, yet the map itself is not used further. \par

In \cite{lee2018implementation} the environment of a telepresence robot is mapped to allow the overlay of virtual fixtures - virtual objects for operator assistance. The motion of the robot arms, however, is guided by the user and no programming was implemented. \par

\subsection{Contributions}
This paper extends the previous approaches in several ways. Firstly by mapping the environment with multi-purpose HMDs we eliminate the need for any overhead equipment for cell setup or the need for CAD data of the surrounding objects. HMDs have been used for robot intention visualisation \cite{Walker2018robotmotion}, collaborative task planning \cite{chakraborti2017} and as previously mentioned programming \cite{Quintero2018ARProg} just to name a few. \par

Secondly the map created this way can be used both for offline programming or as an addition to AR-based approaches such as the one in \cite{Quintero2018ARProg}. In the later case, it allows the use of higher-level motion planning to plan collision-free trajectories, such as MoveIT!. This significantly decreases the programming effort for the user. \par

Finally, we perform thorough tests of the depth sensing capabilities of the HoloLens. As of yet such tests have not been performed. This data will provide useful metrics and possible failure cases for future research. \par  
\section{METHODOLOGY}
\label{sec:methods}

The system consists of two main components, the HoloLens HMD and a desktop computer connected to the robot and running the Robot Operating System (ROS) \cite{ros}. Communication between the desktop and the HoloLens is mediated via the ROSBridge package than allows seamless interfacing between ROS nodes and programs running on different systems. \par

For point cloud editing and filtering the open source Point Cloud Library (PCL) \cite{Rusu_ICRA2011_PCL} was used. The OctoMap representation and the path planning is done using the MoveIt! path planner. \par

On the HoloLens side, the AR interface was constructed using the Unity3D engine. For the capture and streaming of depth information the HoloLensForCV library package was used. Communication with ROS was done using the ROS\# library which provides ROSbridge clients for .Net applications, like Unity3D.\par 


\begin{figure*}[ht!]
\vspace {5pt}
\centering
\subcaptionbox{}
[0.48\textwidth]{ \includegraphics[width=0.48\textwidth]{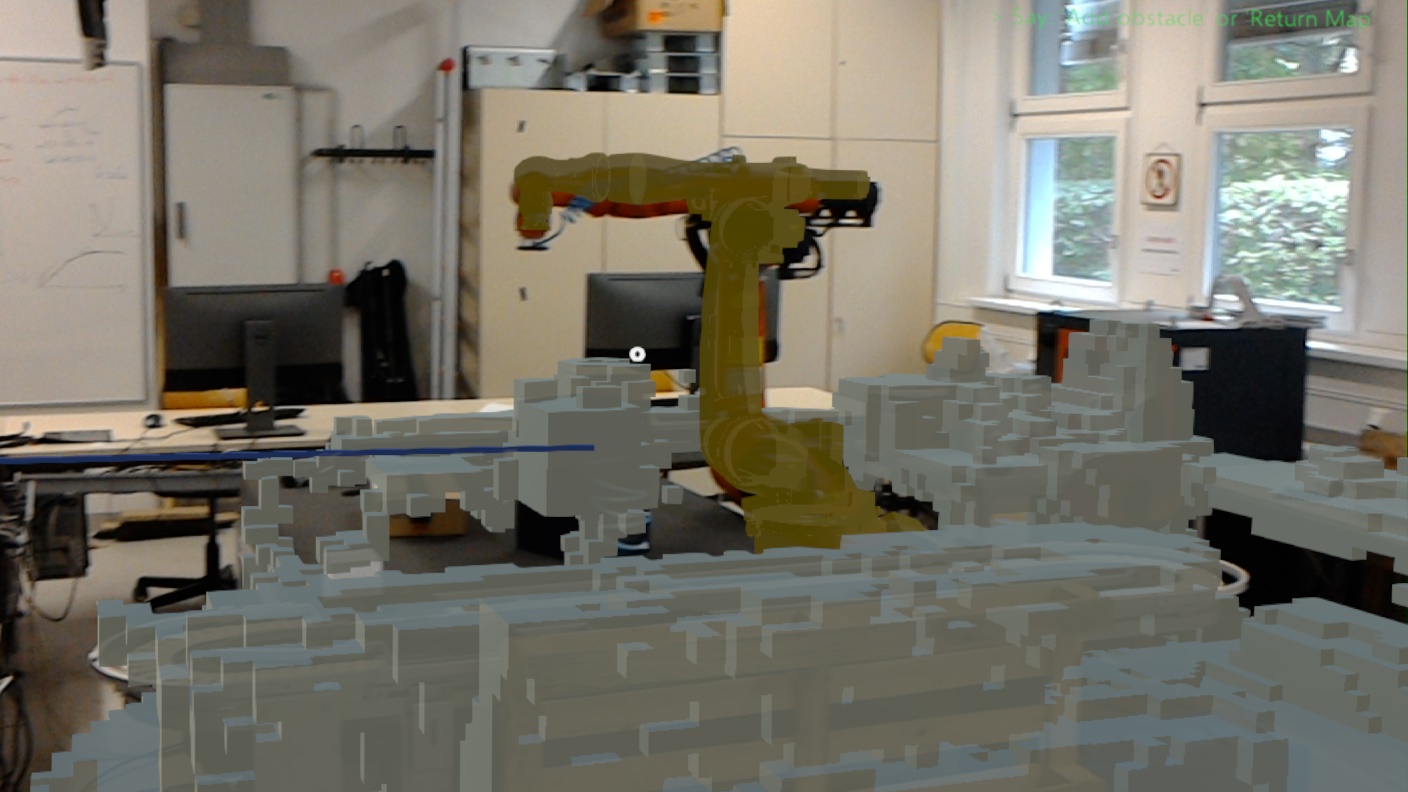}}
\subcaptionbox{}
[0.48\textwidth]{\includegraphics[width=0.48\textwidth]{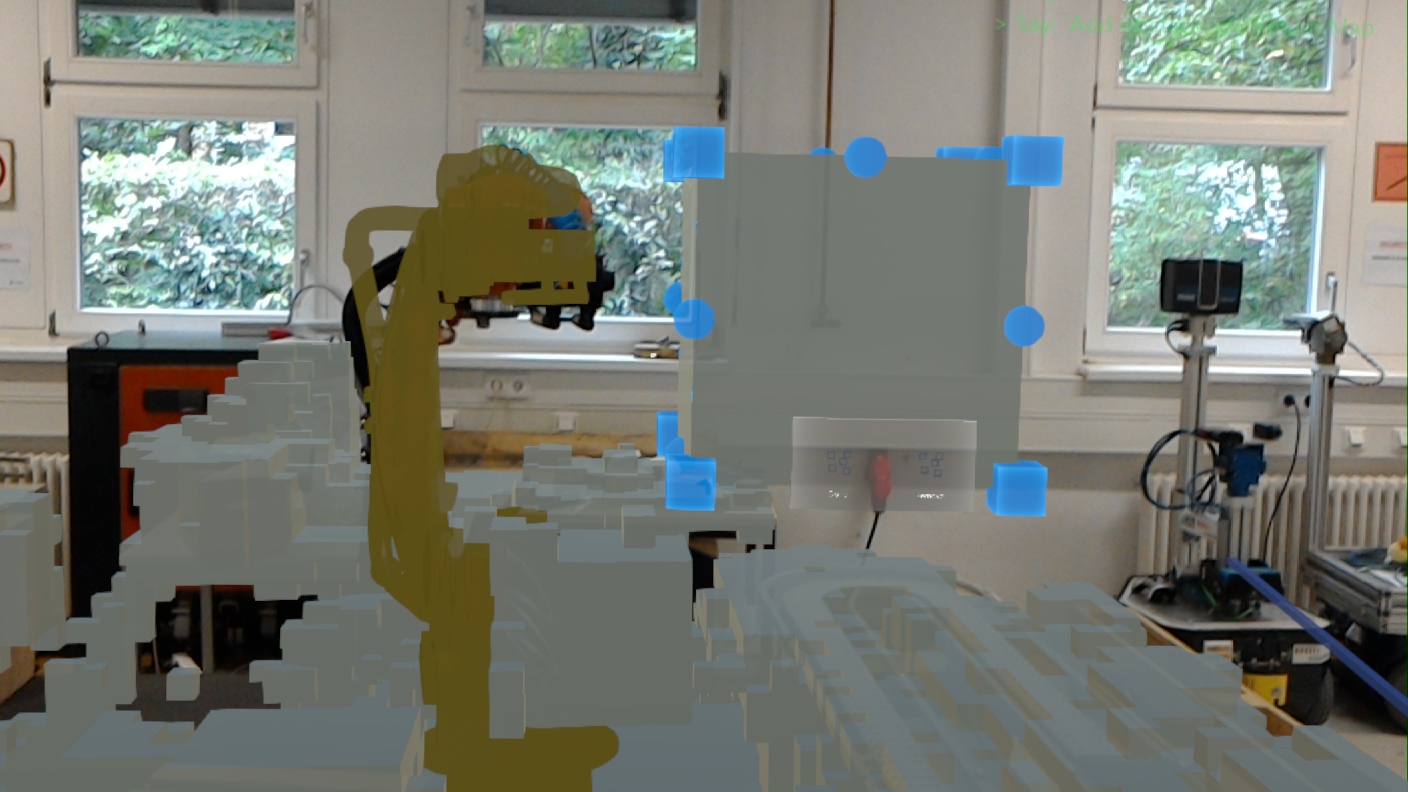}}
    \caption[]{(a) Visualisation of the voxel grid representation of the environment in the HoloLens. The individual voxels can be added, moved and removed; (b) Adding safety zones in situ using the HoloLens. Such definition of safety zones are more intuitive and faster than classical input in offline robot programming.}
    \label{fig:ar}
\end{figure*}

\begin{figure}[hb!]
\includegraphics[width=0.49\textwidth]{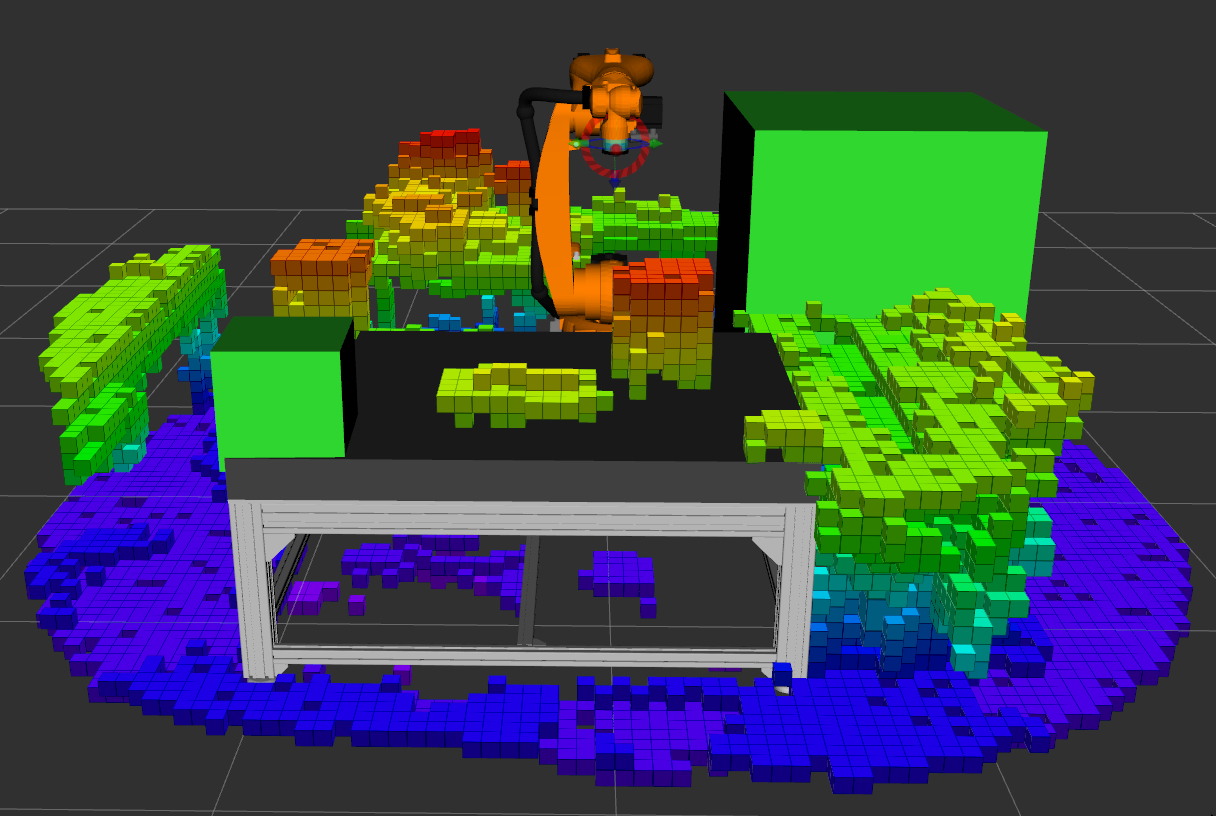}
    \caption[]{The edited environmental voxel grid and safety zones as visualised in RViz. To note is that the table in this application was part of the robot description file. If a CAD model exists and the robot should interact with that part of the environment, adding it to the robot model will filter out the unwanted voxels automatically. }
    \label{fig:rviz}

\end{figure}

\subsection{Mapping}
\label{sec:mapping}
Two different methods of obtaining the point cloud of the environment have been implemented. The first approach uses the mesh of the environment already generated by the HoloLens. Randomly a mesh triangle is chosen, weighted by the size of the triangles. Then, using barycentric coordinates, a random point within the triangle is selected and saved to the point cloud. The number of iterations of
this process, and therefore the size of the resulting point cloud, can be chosen. The resulting point cloud is filtered with voxel grid filtering to obtain a uniform point density. \par

After Microsoft allowed access to the depth stream with the research mode, the raw depth data could be used and the point cloud generated directly. The depth sensor on the HoloLens provides two depth streams, the short-throw depth stream, with 30 frames per second update rate and a range of 0.2-1 meters, and a long-throw depth stream, with 1-5 frames per second update rate and a range of 0.5-4 meters. \par

Combined with the localisation capabilities of the HoloLens, the different depth frames can be fused into a single point cloud of the environment. We registered the point data of different frames to the main point cloud using the HoloLens' own localisation as the initial guess and ICP \cite{icp} to refine the guess. It was found, however, that the HoloLens' localisation is precise enough that ICP does not significantly increase the precision. Therefore the registration step may be skipped. This shall be demonstrated in the experiments in Section \ref{sec:exp_ds}. \par

We then discard points that are below the minimum cut-off distance $D_{cut-off\_min}$ to eliminate points that may belong to the user's hand, and above the maximum cut-off distance $D_{cut-off\_max}$ to eliminate low quality points. The maximum cut-off distance was experimentally determined to be 3.3 meters (Fig.~\ref{fig:cutoff}). The minimum cut-off distance was taken to be one meter, around the reach of the user's arms. Therefore we can use only the long-throw stream and discard points further than 3.3 meters. \par

The resulting point cloud is down-sampled using voxel-grid filtering to ensure uniform point density. It is then filtered with an outlier removal filter, removing any point that had less than 9 neighbours in a radius of 5 cm, and smoothed with moving least squares \cite{Alexa2003MLS}. Finally RANSAC plane detection is used to detect planes and map all the points near the plane to the plane itself. This improves the resolution of objects on floors and tables. \par 

\subsection{Referencing}
\label{sec:referencing}
To get a robust coordinate transform between the HoloLens and the robot world coordinate system, a semi-automatic referencing approach is used \cite{puljiz2019referencing}. The user is asked to position a seed hologram near the robot base and rotate it approximately towards the front of the robot. Using the universal robot description file (urdf) and the link meshes of the robot a point cloud of the robot is created. The model, together with the map of the environment and the position of the seed hologram, are used as the input to an ICP registration algorithm. As the ICP is highly sensitive to local minima, the seed cube is paramount to get a robust coordinate transform. As shown in \cite{puljiz2019referencing} the positioning of the seed hologram doesn't have to be precise but merely near the base of the robot. \par      

\subsection{Workspace Representation and Editing}
\label{sec:octomap}

After obtaining the position of the robot in the HoloLens' environmental point cloud in the referencing step, all points of the map outside the maximum reach of the robot are removed. To do this a kD-tree representation of the point cloud is first constructed. The kD-tree is a data structure that facilitates radius and nearest-neighbour searches. We then filter all the points further away than $D_{reach}$ from the (0,0,0) point, which is taken as the base of the robot. \par

The point cloud is then once again down-sampled with a voxel-grid filter with voxel size $D_{leaf}$, which is the size of the voxels in the OctoMap. For each point in the model point cloud of the robot, a nearest-neighbour kD-tree search is performed to remove all the robot points from the scene. Finally the resulting point cloud is used to generate an OctoMap voxel grid representation of the occupancy as seen in Fig.~\ref{fig:intro}. \par

The generated voxel occupancy grid is sent to the HoloLens where a user may edit the occupancy grid. This step allows the user to correct errors in the map if needed be. It also allows for voxel removal from parts of the environment where contact with the environment is needed to perform the robot's task. Finally, safety zones can be defined in situ, drastically reducing the set-up and test times. In Fig.~\ref{fig:ar} the overlayed robot model, the rendered OctoMap, and the set up of the safety zones can be seen.  \par

When the user is done, the safety zones and the edited OctoMap are sent back to the computer. As the user can freely move and add voxels through AR on the HoloLens, these voxels must be snapped back to the voxel grid. These OctoMap environmental representations can be saved, loaded in MoveIt! and edited with the HoloLens as many times as necessary. Likewise, one could safe different voxel grids and safety zones depending on the task for future uses. A representation of an edited map and safety zones in RViz can be seen in Fig.~\ref{fig:rviz}.

\section{EXPERIMENTS AND RESULTS}
\label{sec:experiments}
In this section we present in-depth tests of the HoloLens' spatial mapping capabilities. The experiments to test the precision of the direct measurements from the depth sensor as well as the quality of the resulting point cloud are presented in Section \ref{sec:exp_ds}. In Section \ref{sec:exp_om} we present the experiments and the results to test the quality of the resulting OctoMap by counting the amount of false-positive and false-negative voxels in the occupancy grid of a test scene. Further tests with the robot and the motion planer itself revealed a particular failure case which will be addressed in Section \ref{sec:exp_disc}. The results of the previous tests shall also be discussed. \par     


\subsection{Evaluation of Depth Sensing Capabilities}
\label{sec:exp_ds}
\begin{figure}[hb]
\centering
\subcaptionbox{}
[0.22\textwidth]{ \includegraphics[width=0.22\textwidth]{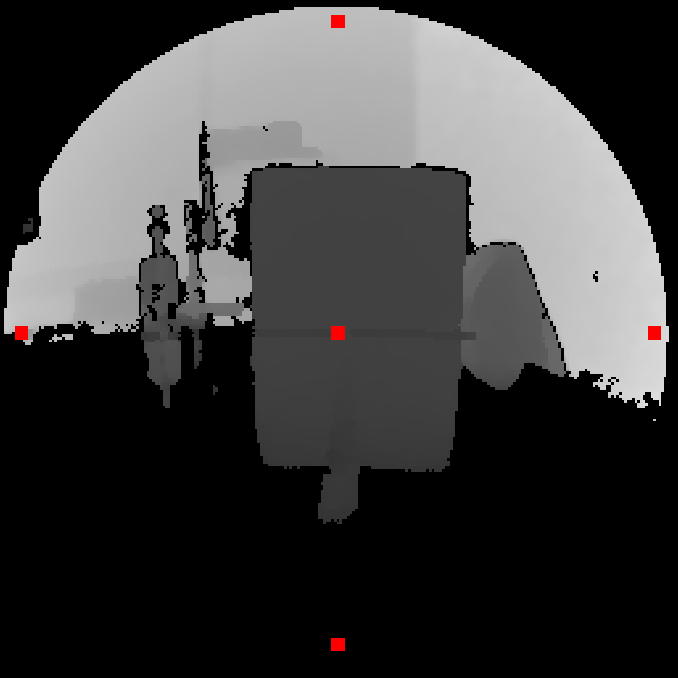}}
\subcaptionbox{}
[0.22\textwidth]{\includegraphics[width=0.22\textwidth]{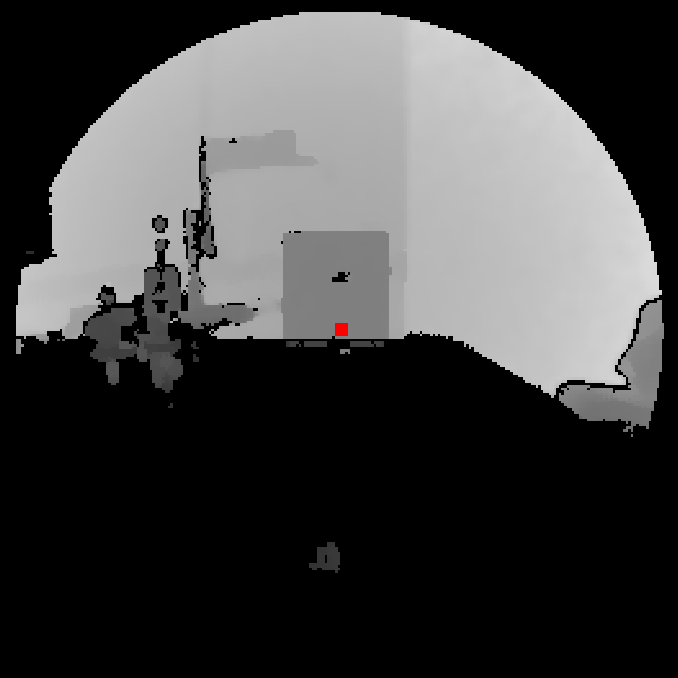}}
    \caption[]{(a) Experiment 1 - A flat cardboard surface was placed 1 meter from the HoloLens' depth sensor. The HoloLens was oriented such that each of the 5 points is on the cardboard surface. It was repeated for 1 and 2 meters respectively. (b) Experiment 2 - a 5x5 pixel area in the centre was taken and the average distance and the standard deviation inside the area were calculated. Again the experiment was repeated at 1 and 2 meters.}
    \label{fig:depth_sens}
\end{figure}

The first set of experiments were aimed to test the noise of the depth sensor data as well as any influence of the position of the pixel. A flat cardboard surface was positioned at 1 and 2 meters respectively from the HoloLens' depth sensor. The HoloLens was rotated so that one of the designated five points pictured in Fig.~\ref{fig:depth_sens}(a) lies on the cardboard surface. A total of fifteen consecutive depth frames were taken for each pixel and each distance for a total of 150 measurements. The results are depicted in Table ~\ref{tab:exp1}. The standard deviation of the depth measurement fluctuations around the average was found to be 3 mm and the maximum fluctuation around the average 5 mm.  \par

\begin{table}[H]
\caption{The distances with smallest and highest error as well as the average distance and standard distance deviations for one and two meters respectively. Measured in meters.}
\label{tab:exp1}
\centering
\resizebox{\columnwidth}{!}{
 \begin{tabular}{| c | c | c | c | c | c |}
 	\hline
 	& Center & Top & Right & Bottom & Left \\ \hline
 	Minimum Error Distance 1m $[m]$ & $1.050$ & $1.016$ & $1.012$ & $0.885$ & $1.016$ \\ \hline
 	Maximum Error Distance 1m $[m]$ & $1.056$ & $1.024$ & $1.015$ & $0.893$ & $1.020$ \\ \hline
 	Average Distance 1m $[m]$ & $1.05233$ & $1.01907$ & $1.0138$ & $0.88813$ & $1.018$ \\ \hline
 	Standard Deviation Distance 1m $[m]$ & $0.00171$ & $0.00228$ & $0.00063$ & $0.00269$ & $0.00106$ \\ \hline
 	Minimum Error Distance 2m $[m]$ & $2.006$ & $2.005$ & $2.001$ & $2.042$ & $2.004$ \\ \hline
 	Maximum Error Distance 2m $[m]$ & $2.013$ & $2.012$ & $2.006$ & $2.049$ & $2.008$ \\ \hline
 	Average Distance 2m $[m]$ & $2.00907$ & $2.00907$ & $2.00387$ & $2.0458$ & $2.00567$ \\ \hline
 	Standard Deviation Distance 2m $[m]$ & $0.00200$ & $0.00222$ & $0.00130$ & $0.00231$ & $0.00100$ \\ \hline
 \end{tabular}
 }
\end{table}

In the second set of experiments a 5x5 pixel square in the centre of the depth image was selected and the values measured. Again a flat cardboard surface was placed 1 and 2 meters away respectively. For each distance 5 repetitions were carried out to average out human positioning error. For each repetition 5 consecutive frames were used for a total of 50 measurements. The setup can be seen in Fig.~\ref{fig:depth_sens}(b). The averages can be seen in Table ~\ref{tab:exp2}. The maximum error of the averages of each square is 11.2 mm and the total average error is 6 mm. \par

\begin{figure*}[ht]
\vspace {5pt}
\centering
\includegraphics[width=0.98\textwidth]{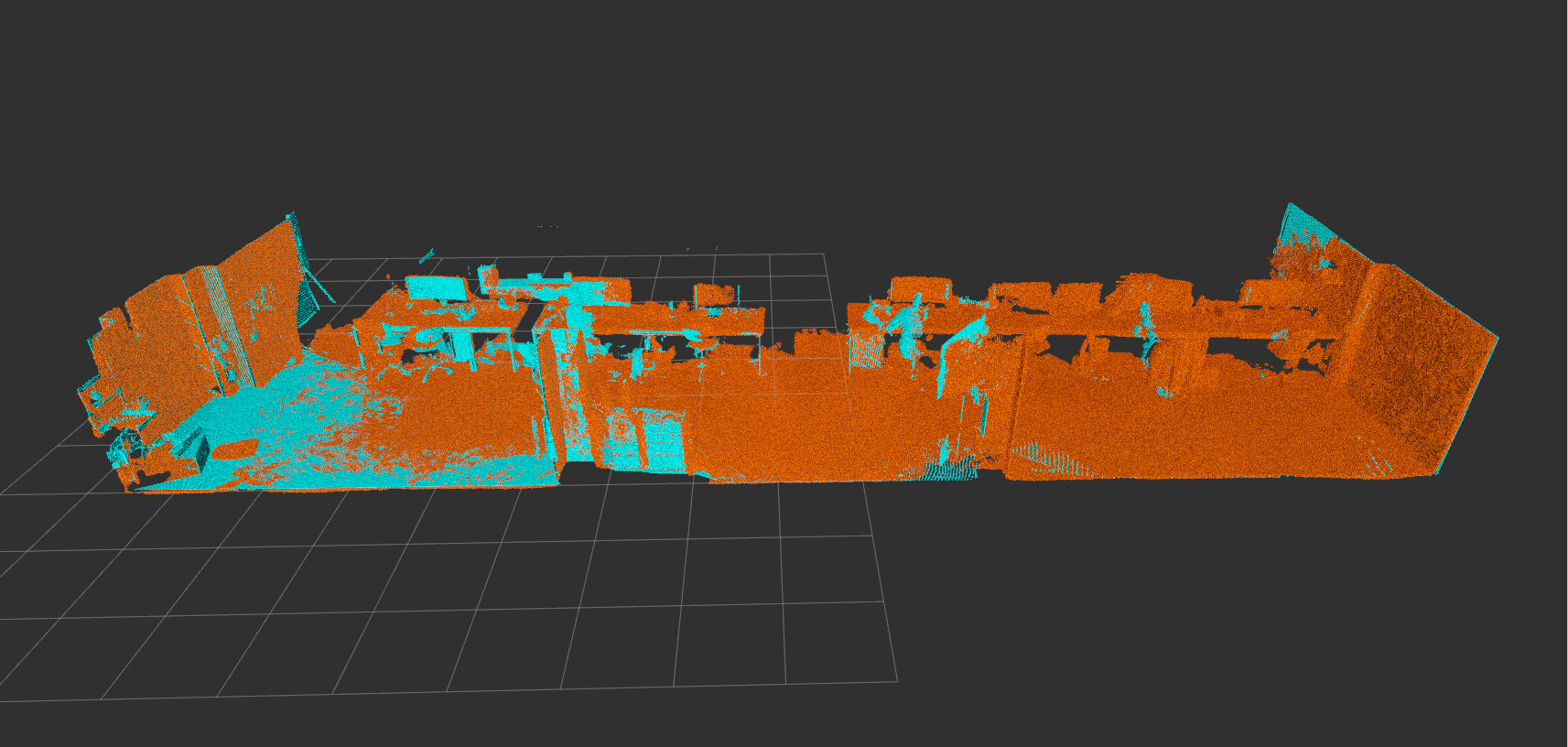}
    \caption[]{The overlapping segment of the laser scan - cyan and the HoloLens point cloud - orange used to calculate the average euclidean distance and the Hausdorff distance in experiment 3.}
    \label{fig:scan}
\end{figure*}

\begin{table}[H]
\caption{The observed averaged depth values for each  repetition of experiment 2.}
\label{tab:exp2}
\centering
\resizebox{\columnwidth}{!}{
 \begin{tabular}{| c | c | c | c | c | c|}
 	\hline
 	{} & Repetition 1 & Repetition 2 & Repetition 3 & Repetition 4 & Repetition 5 \\ \hline \hline
 	1 Meter & $0.998400$  & $1.007512$  & $0.998728$ & $1.005440$ &$1.003424$ \\ \hline 
 	2 Meters & $2.011248$  & $2.006576$  & $2.005984$ & $2.008064$  & $2.000224$ \\ \hline
 \end{tabular}
 }
\end{table}

In the third set of experiments we took scans of our laboratory using a Faro Foucus\textsuperscript{S} laser scanner with 1 mm precision as the ground truth. We compared it to a point cloud generated by the HoloLens. The HoloLens' point cloud was tested with four different combinations of using ICP for registration or not and using the post-processing step, consisting of MLS smoothing and RANSAC plane detection and projection, or not. Firstly we selected the parts of the environment where the two scans overlap (see Fig.~\ref{fig:scan}) and measured the average euclidean distances and the Hausdorff distances, the greatest distance between two closest points in the two point clouds, of the four combinations. The results are presented in Table ~\ref{tab:exp3}. One can see that apparently the post-processing step introduces a bigger error. The large Hausdorff distance can likewise be attributed to left-over discrepancies in the two point clouds either as a result of missed holes or the fact that the point clouds are taken at slightly different time in a changing environment. \par

\begin{table}[H]
\caption{The average euclidean distances and the Hausdorff distance between the laser scan, ground truth point cloud and the HoloLens' point cloud}
\label{tab:exp3}
\centering
\resizebox{\columnwidth}{!}{
 \begin{tabular}{| c | c | c | c | c |}
 	\hline
 	\multirow{2}{*} & Without ICP, & With ICP, & Without ICP, & With ICP, \\
 	& not postprocessed & not postprocessed & postprocessed & postprocessed \\ \hline
 	Euclidean distance $[m]$ & $0.040128$ & $0.040742$ & $0.061583$ & $0.062344$ \\ \hline
 	Hausdorff distance $[m]$ & $1.052818$ & $1.054748$ & $1.172284$ & $1.169257$ \\ \hline
 \end{tabular}
 }
\end{table}

\begin{figure*}[ht!]
\vspace {5pt}
\centering
\subcaptionbox{}
[0.48\textwidth]{ \includegraphics[width=0.48\textwidth]{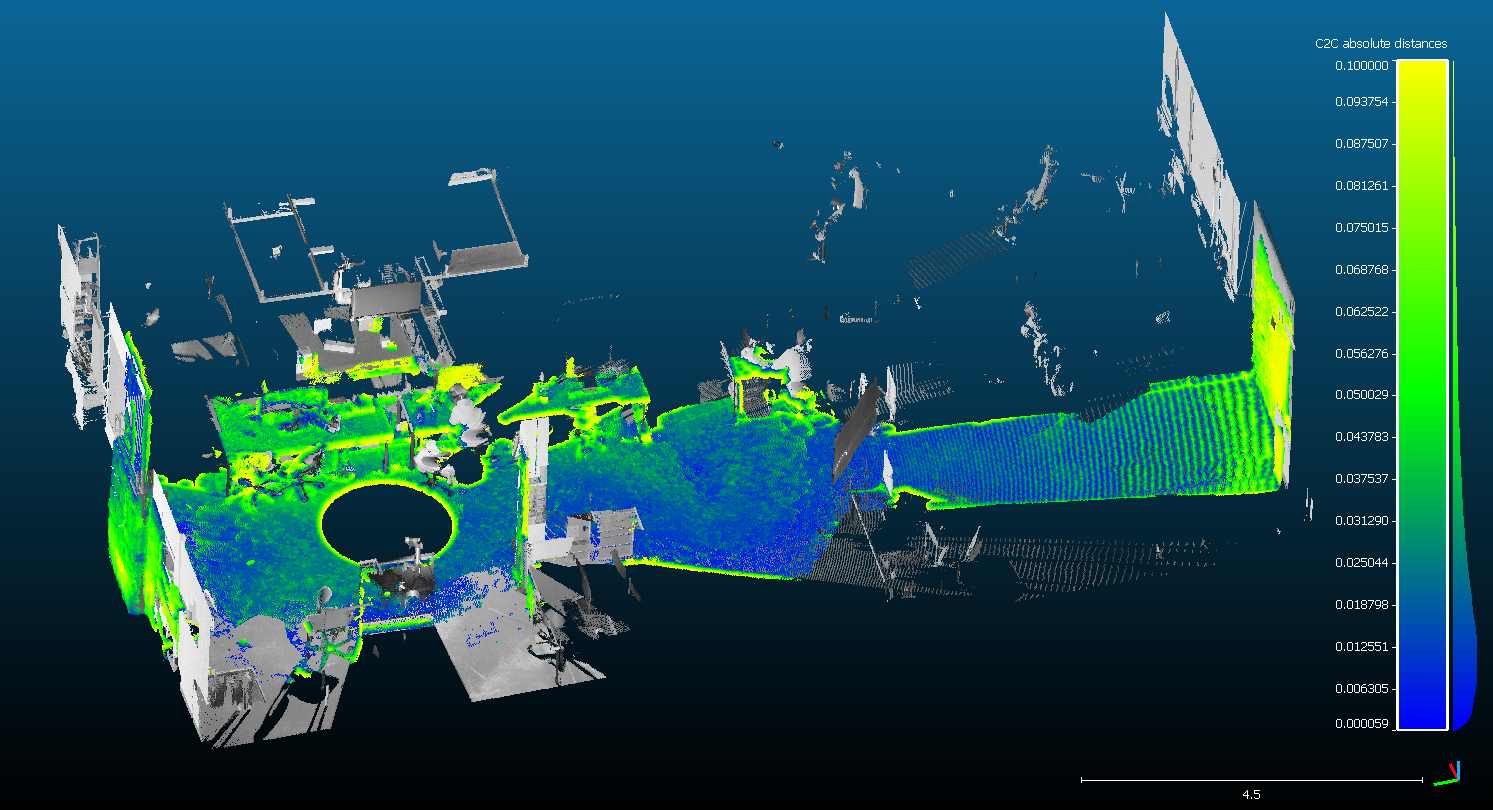}}
\subcaptionbox{}
[0.48\textwidth]{\includegraphics[width=0.48\textwidth]{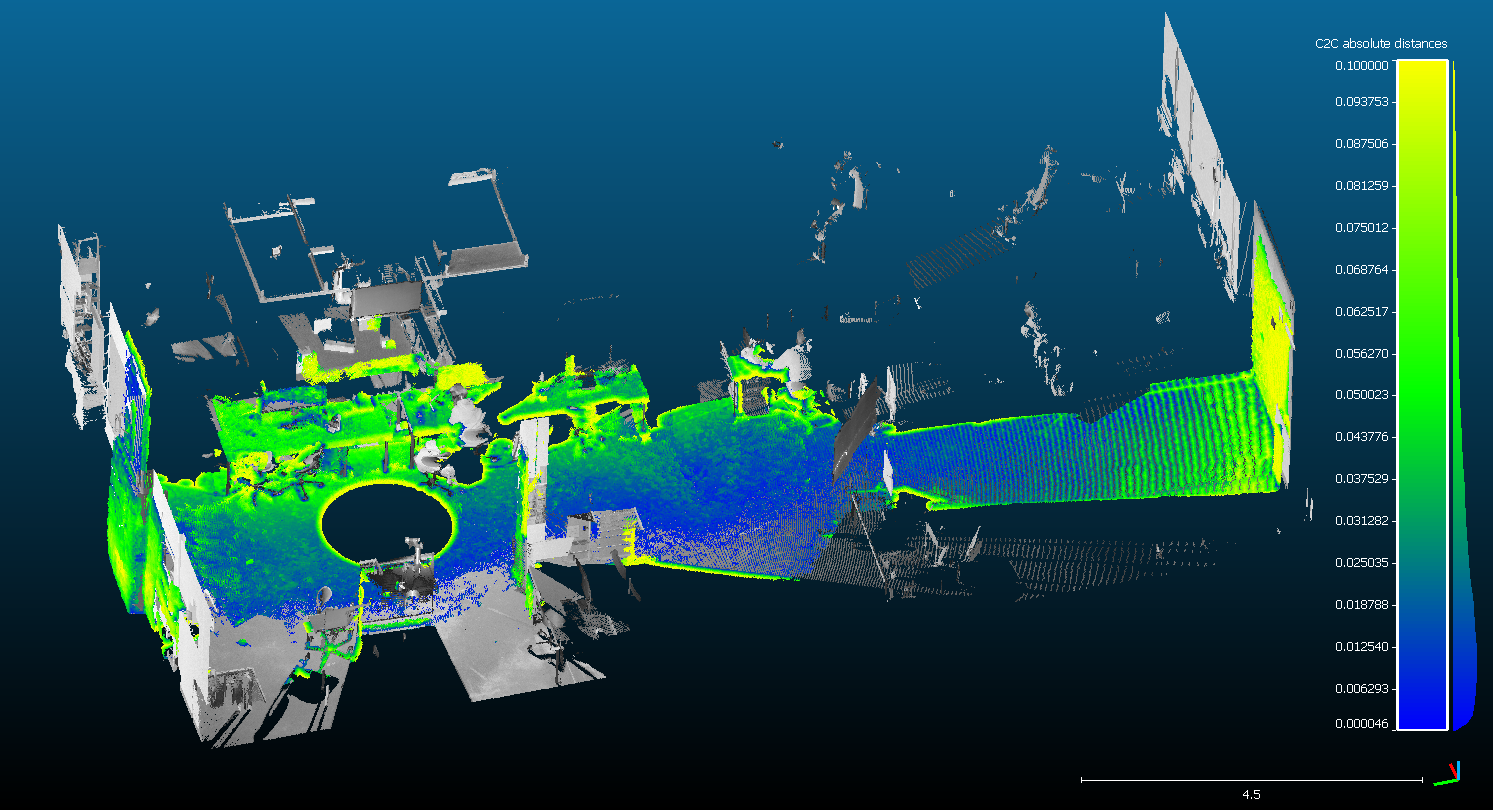}}
\subcaptionbox{}
[0.48\textwidth]{\includegraphics[width=0.48\textwidth]{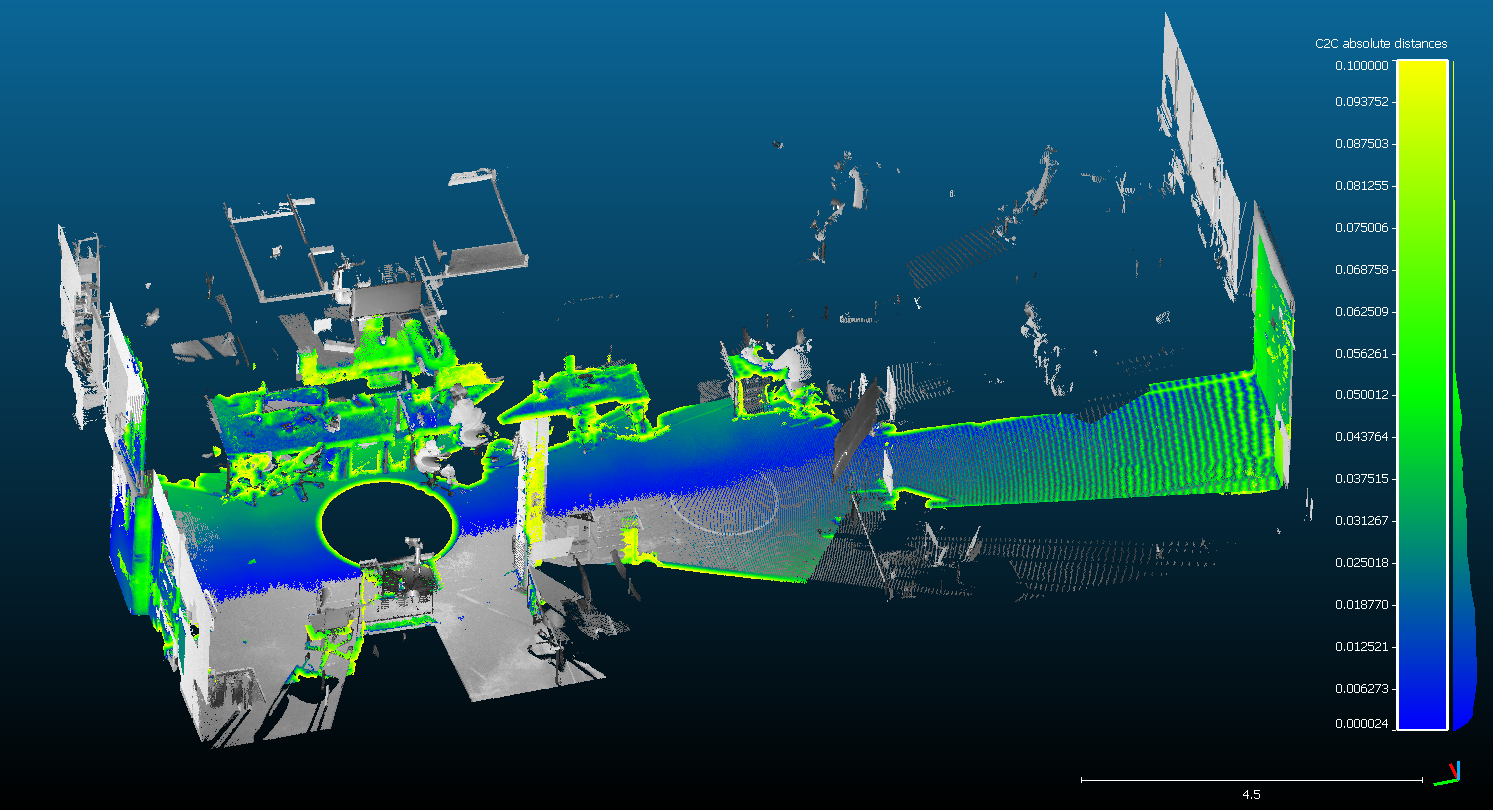}}
\subcaptionbox{}
[0.48\textwidth]{\includegraphics[width=0.48\textwidth]{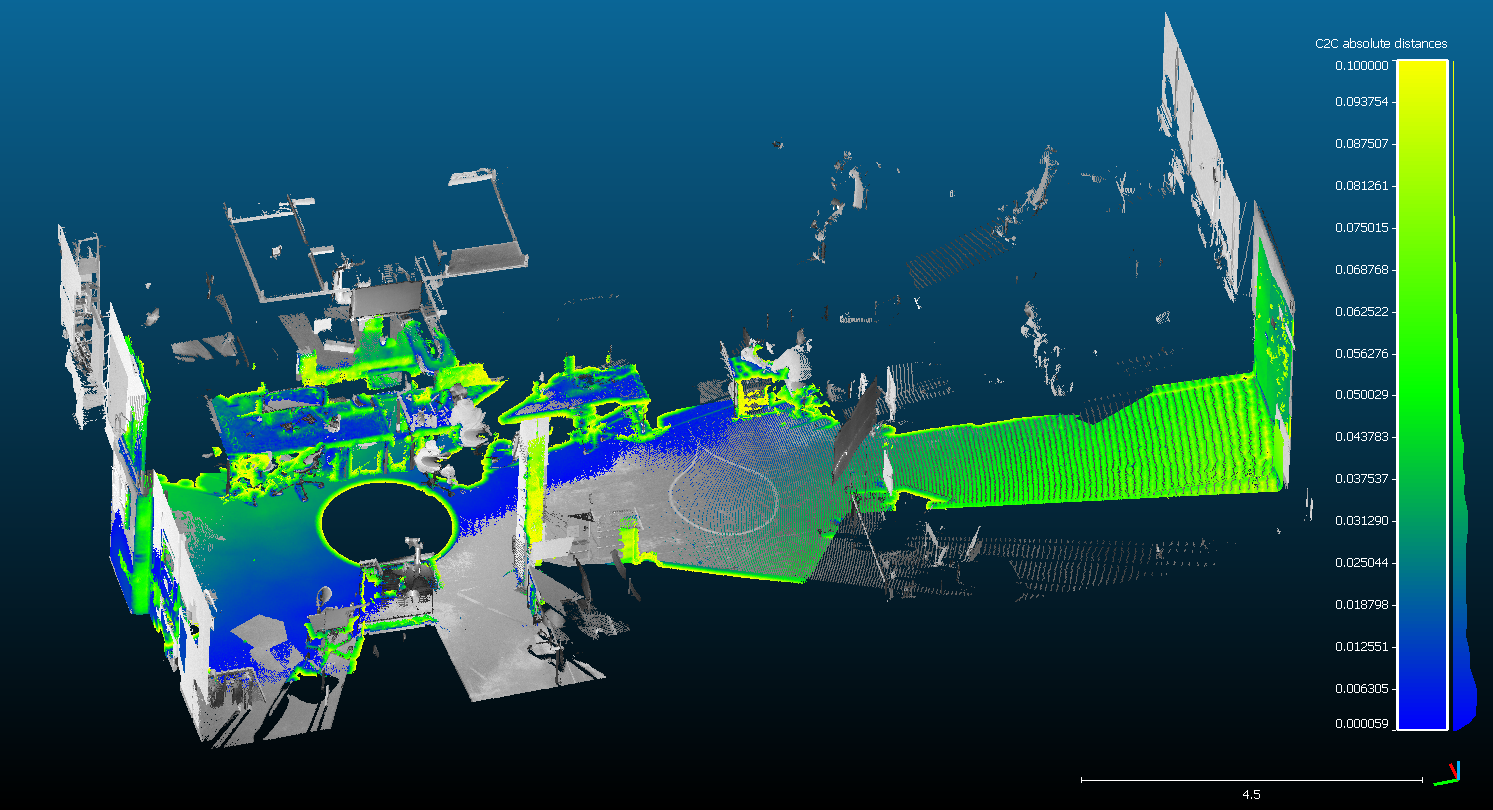}}    \caption[]{Comparison of the HoloLens point cloud with the ground truth obtained via laser scan;(a) Point cloud without ICP registration between frames and without the post-processing step of MLS smoothing and RANSAC plane detection and projection;(b) Point cloud with ICP registration and without postprocessing; (c) Point cloud without ICP registration but with post processing; (d) Point cloud with ICP registration and with post-processing.}
    \label{fig:cloudcompare}
\end{figure*}

To get a better estimate of the precision of the two point clouds we used CloudCompare. CloudCompare gives the percentile distribution of distances between the two point clouds and therefore offers a much better insight into the quality of the point cloud generated from the HoloLens. The results are shown in Table ~\ref{tab:exp4}. One can see that the best performance is the mapping without ICP, meaning that the HoloLens localisation is as precise as the point cloud, and with post-processing. In this case 75 percent of points have an error of 3.6 cm or lower. A visual comparison of the four point clouds to the ground truth can be seen in Fig.~\ref{fig:cloudcompare}.

\begin{table}[H]
\caption{The percentiles of the distances of each spatial map combination to the laser scan.}
\label{tab:exp4}
\centering
\resizebox{\columnwidth}{!}{
 \begin{tabular}{| c | c | c | c | c |}
 	\hline
 	\multirow{2}{*}{Percentile} & Without ICP, & With ICP, & Without ICP, & With ICP, \\
 	& not postprocessed & not postprocessed & postprocessed & postprocessed \\ \hline
 	10th $[m]$ & $0.00591$ & $0.00629$ & $0.00588$ & $0.00513$ \\ \hline
 	25th $[m]$ & $0.01177$ & $0.01254$ & $0.01135$ & $0.01099$ \\ \hline
 	50th $[m]$ & $0.02192$ & $0.02503$ & $0.02150$ & $0.02582$ \\ \hline
 	75th $[m]$ & $0.04105$ & $0.04573$ & $0.03673$ & $0.04183$ \\ \hline
 	90th $[m]$ & $0.06447$ & $0.06994$ & $0.05275$ & $0.06057$ \\ \hline
 \end{tabular}
 }
\end{table}

\subsection{Evaluation of the OctoMap}
\label{sec:exp_om}

To evaluate the quality of the OctoMap generated from the HoloLens' point cloud we placed a wooden 20x20x30 cm cuboid on the front-left of the table. We assumed the worst case scenario of using the sampled environmental mesh of the HoloLens. Khoshelham et al. \cite{Khoshelham2019SpatialMapping} found that the average global error of the HoloLens' environmental mesh is around 5 cm. We counted the total number of false-positives, i.e. the voxels that are detected as occupied by the object that are in fact not, and false-negatives, i.e. voxels detected as free that are in fact part of the object. The results presented in Table ~\ref{tab:exp_oct} show that on average there are 9.58 false-negatives and 61.33 false-positives with 12 point clouds tested. Worth noting is that the false-negatives are much more critical as they can cause crashes while false-positives only slightly limit the collision-free volume. Also worth noting is that some false-negatives are hidden behind false-positive voxels or near the table and are therefore unreachable. \par

We also carried more than 50 tests to try to provoke collision in the cluttered scene shown in Fig.~\ref{fig:intro}. The tests showed that there are indeed edge cases were a collision might happen i.e. when objects are positioned diagonally. A solution for these edge cases are presented in the next subsection. \par    

\subsection{Discussion}
\label{sec:exp_disc}

We have shown that the HoloLens environment mapping capabilities, with a MLS smoothing and RANSAC plane finding and fitting stage can produce a point cloud where 75 percent of the point lie within 3.6 cm of a ground truth point cloud captured with a high-end laser scanner. \par  
The OctoMap voxel occupancy grid performs adequately even in the worst case scenario where a sampled environemtnal mesh of the HoloLens was used. \par

\begin{figure*}[ht]
\vspace {5pt}
\centering
\subcaptionbox{}
[0.32\textwidth]{ \includegraphics[width=0.32\textwidth]{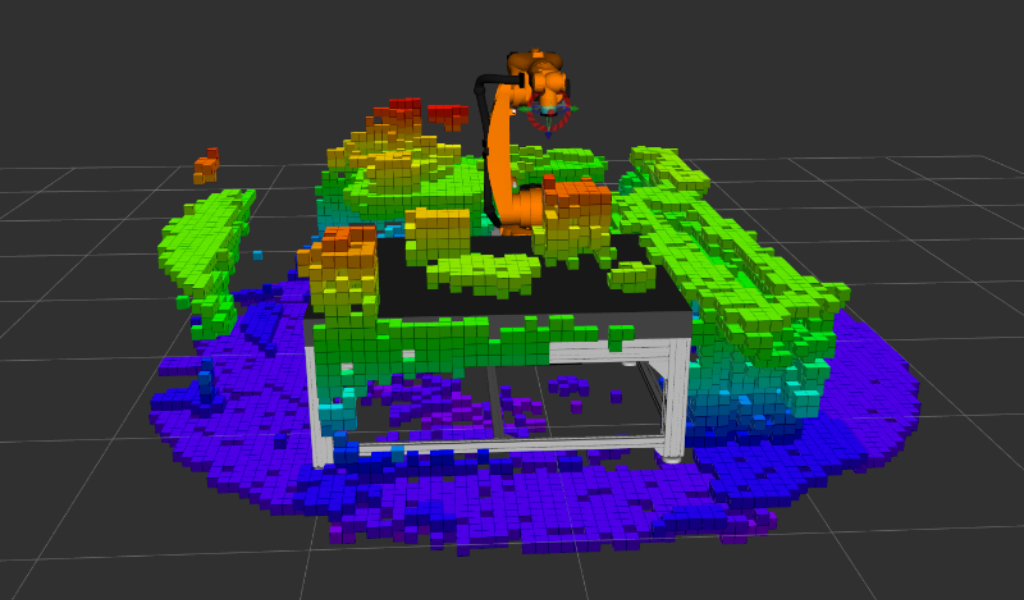}}
\subcaptionbox{}
[0.32\textwidth]{\includegraphics[width=0.32\textwidth]{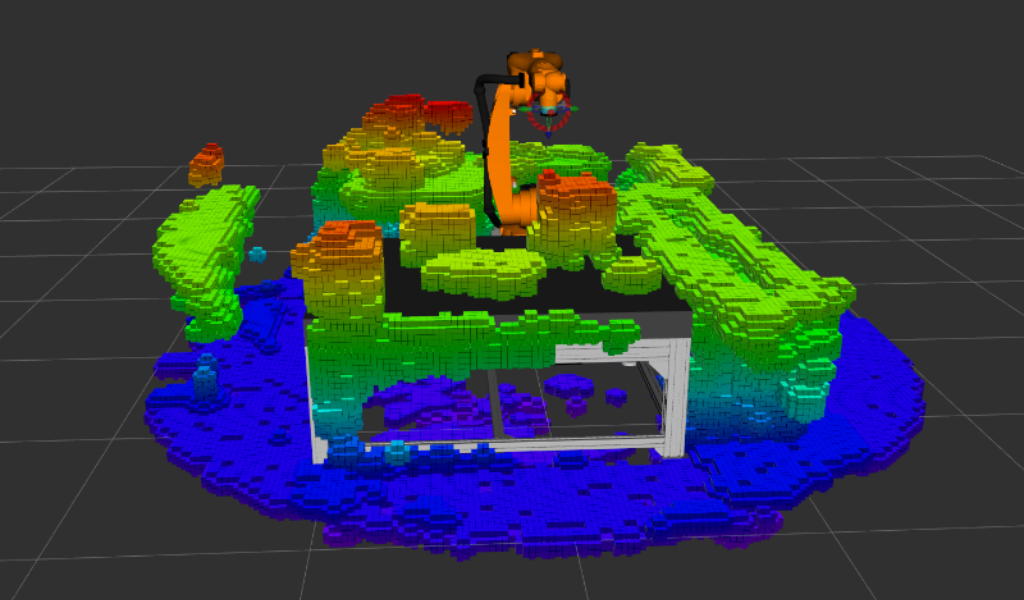}}
\subcaptionbox{}
[0.32\textwidth]{\includegraphics[width=0.32\textwidth]{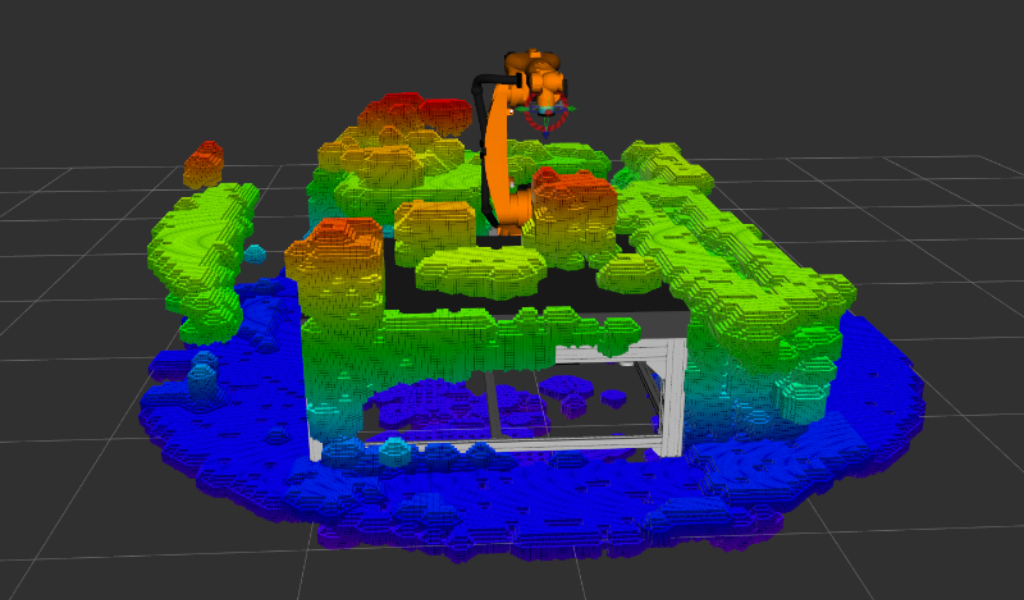}}

    \caption[]{The padding process to remove the edge cases that might result in collisions. (a) The original OctoMap; (b) First level padding with voxels of size 2.5 cm; (c) second level padding with voxels of size 1.25 cm on top of the level 1 padding.}
    \label{fig:padding}
\end{figure*}

To mitigate the edge cases, a padding algorithm was developed. Each new level pads the surface of the starting voxels iteratively with voxels half the size of the starting voxels, as illustrated in Fig.~\ref{fig:padding}. Even with one level of padding, it was shown that the edge cases were eliminated and no collisions occurred anymore.  

\begin{table*}[h!]
\caption{The number of false-positives and false-negatives in the 12 OctoMaps tested}
\label{tab:exp_oct}
\centering
\resizebox{2\columnwidth}{!}{
\begin{tabular}{|l||l|l|l|l||l|l|l|l||l|l|l|l||c|c|}
\hline
& \textbf{1}  & \textbf{2}  & \textbf{3}  & \textbf{4}  & \textbf{5}  & \textbf{6}  & \textbf{7}  & \textbf{8}  & \textbf{9}  &\textbf{10} & \textbf{11} & \textbf{12} & \textbf{$\varnothing$} & \textbf{$\sigma$}\\ \hline \hline
false positive & 83 & 60 & 67 & 71 & 52 & 75 & 78 & 61 & 41 & 35 & 58 & 55 & 61.33 & 14.48  \\ \hline
false negative & 6  & 16 & 10 & 5  & 5  & 8  & 14 & 7  & 12 & 15 & 11 & 6  & 9.58 &  3.99   \\ \hline
\end{tabular}
}
\centering
\end{table*}

\section{CONCLUSION AND FUTURE WORK}
\label{sec:conclusion}
For robotic applications to really become ubiquitous in enterprises of any size, easy set-up and programming of robots is crucial. In this paper we presented a robot cell modelling approach that relies on the Microsoft HoloLens to reference and map the previously unknown environment of the robot. As the amount of research in AR-based human-robot interaction and programming is has seen major growth in recent years, a plethora of such programs and research can be combined with our approach to reduce user workload and extend the area of possible applications. \par

We have proven that the mapping and localisation capabilities of the HoloLens are more than adequate for such an application. Even if errors do occur, the interactive editing of the environmental map and the safety zones can quickly remove such errors. Furthermore the map can be edited for different task quickly and efficiently. \par

There is still plenty of room for improvement, however. We developed a programming approach based on the work of Quintero et al. \cite{Quintero2018ARProg}, which should be integrated with our approach. User tests should be made to see if the perceived workload of the users is indeed lowered when using the proposed environmental mapping. Another exciting research direction is using a dynamic map for the interaction with the robot as well as sharing and fusing of sensor data if the robot is also equipped with sensors. In the authors opinion such on-line sharing of data could be a great benefit in proximal human-robot collaboration. \par  

\section*{ACKNOWLEDGMENT}
This work has been supported from the European Union’s Horizon 2020 research and innovation programme under grant agreement No 688117 “Safe human-robot interaction in logistic applications for highly flexible warehouses (SafeLog)”.


\bibliographystyle{IEEEtran}
\bibliography{bib}

\begin{thebibliography}{10}
\providecommand{\url}[1]{#1}
\csname url@samestyle\endcsname
\providecommand{\newblock}{\relax}
\providecommand{\bibinfo}[2]{#2}
\providecommand{\BIBentrySTDinterwordspacing}{\spaceskip=0pt\relax}
\providecommand{\BIBentryALTinterwordstretchfactor}{4}
\providecommand{\BIBentryALTinterwordspacing}{\spaceskip=\fontdimen2\font plus
\BIBentryALTinterwordstretchfactor\fontdimen3\font minus
  \fontdimen4\font\relax}
\providecommand{\BIBforeignlanguage}[2]{{%
\expandafter\ifx\csname l@#1\endcsname\relax
\typeout{** WARNING: IEEEtran.bst: No hyphenation pattern has been}%
\typeout{** loaded for the language `#1'. Using the pattern for}%
\typeout{** the default language instead.}%
\else
\language=\csname l@#1\endcsname
\fi
#2}}
\providecommand{\BIBdecl}{\relax}
\BIBdecl

\bibitem{Pan2012ProgrammingMethods}
Z.~{Pan}, J.~{Polden}, N.~{Larkin}, S.~V. {Duin}, and J.~{Norrish}, ``Recent
  progress on programming methods for industrial robots,'' in \emph{ISR 2010
  (41st International Symposium on Robotics) and ROBOTIK 2010 (6th German
  Conference on Robotics)}, June 2010, pp. 1--8.

\bibitem{schraft2006need}
R.~D. Schraft and C.~Meyer, ``The need for an intuitive teaching method for
  small and medium enterprises,'' \emph{VDI BERICHTE}, vol. 1956, p.~95, 2006.

\bibitem{Quintero2018ARProg}
C.~P. Quintero, S.~Li, M.~K. Pan, W.~P. Chan, H.~F. M.~V. der Loos, and
  E.~Croft, ``Robot programming through augmented trajectories in augmented
  reality,'' in \emph{2018 IEEE/RSJ International Conference on Intelligent
  Robots and Systems (IROS)}, Oct 2018, pp. 1838--1844.

\bibitem{NETO2013DirectOffLine}
\BIBentryALTinterwordspacing
P.~Neto and N.~Mendes, ``Direct off-line robot programming via a common cad
  package,'' \emph{Robotics and Autonomous Systems}, vol.~61, no.~8, pp. 896 --
  910, 2013. [Online]. Available:
  \url{http://www.sciencedirect.com/science/article/pii/S0921889013000419}
\BIBentrySTDinterwordspacing

\bibitem{Vincze2003Detection}
M.~{Vincze}, A.~{Pichler}, and G.~{Biegelbauer}, ``Detection of classes of
  features for automated robot programming,'' in \emph{2003 IEEE International
  Conference on Robotics and Automation (Cat. No.03CH37422)}, vol.~1, Sep.
  2003, pp. 151--156 vol.1.

\bibitem{ong2010novel}
\BIBentryALTinterwordspacing
S.~K. Ong, J.~W.~S. Chong, and A.~Y. Nee, ``A novel ar-based robot programming
  and path planning methodology,'' \emph{Robotics and Computer-Integrated
  Manufacturing}, vol.~26, no.~3, pp. 240--249, 2010. [Online]. Available:
  \url{{https://www.sciencedirect.com/science/article/abs/pii/S0736584509001100}}
\BIBentrySTDinterwordspacing

\bibitem{lee2018implementation}
\BIBentryALTinterwordspacing
D.~Lee and Y.~S. Park, ``Implementation of augmented teleoperation system based
  on {Robot Operating System (ROS)},'' in \emph{2018 IEEE/RSJ International
  Conference on Intelligent Robots and Systems (IROS)}.\hskip 1em plus 0.5em
  minus 0.4em\relax IEEE, 2018, pp. 5497--5502. [Online]. Available:
  \url{{https://ieeexplore.ieee.org/abstract/document/8594482}}
\BIBentrySTDinterwordspacing

\bibitem{Walker2018robotmotion}
\BIBentryALTinterwordspacing
M.~Walker, H.~Hedayati, J.~Lee, and D.~Szafir, ``Communicating robot motion
  intent with augmented reality,'' in \emph{Proceedings of the 2018 ACM/IEEE
  International Conference on Human-Robot Interaction}, ser. HRI '18.\hskip 1em
  plus 0.5em minus 0.4em\relax New York, NY, USA: ACM, 2018, pp. 316--324.
  [Online]. Available: \url{http://doi.acm.org/10.1145/3171221.3171253}
\BIBentrySTDinterwordspacing

\bibitem{chakraborti2017}
\BIBentryALTinterwordspacing
T.~Chakraborti, S.~Sreedharan, A.~Kulkarni, and S.~Kambhampati, ``Alternative
  modes of interaction in proximal human-in-the-loop operation of robots,''
  \emph{CoRR}, vol. abs/1703.08930, 2017. [Online]. Available:
  \url{http://arxiv.org/abs/1703.08930}
\BIBentrySTDinterwordspacing

\bibitem{ros}
M.~Quigley, B.~Gerkey, K.~Conley, J.~Faust, T.~Foote, J.~Leibs, E.~Berger,
  R.~Wheeler, and A.~Ng, ``Ros: an open-source robot operating system,'' in
  \emph{Proc. of the IEEE Intl. Conf. on Robotics and Automation (ICRA)
  Workshop on Open Source Robotics}, Kobe, Japan, May 2009.

\bibitem{Rusu_ICRA2011_PCL}
R.~B. Rusu and S.~Cousins, ``{3D is here: Point Cloud Library (PCL)},'' in
  \emph{{IEEE International Conference on Robotics and Automation (ICRA)}},
  Shanghai, China, May 9-13 2011.

\bibitem{icp}
P.~J. Besl and N.~D. McKay, ``A method for registration of 3-d shapes,''
  \emph{IEEE Transactions on Pattern Analysis and Machine Intelligence},
  vol.~14, no.~2, pp. 239--256, Feb 1992.

\bibitem{Alexa2003MLS}
M.~{Alexa}, J.~{Behr}, D.~{Cohen-Or}, S.~{Fleishman}, D.~{Levin}, and C.~T.
  {Silva}, ``Computing and rendering point set surfaces,'' \emph{IEEE
  Transactions on Visualization and Computer Graphics}, vol.~9, no.~1, pp.
  3--15, Jan 2003.

\bibitem{puljiz2019referencing}
\BIBentryALTinterwordspacing
D.~Puljiz, K.~S. Riesterer, B.~Hein, and T.~Kr{\"o}ger, ``Referencing between a
  head-mounted device and robotic manipulators,'' in \emph{Proceedings of the
  2nd Workshop on Virtual, Mixed and Augmented Reality Human.Robot Interaction,
  HRI 2019}, 2019. [Online]. Available: \url{http://arxiv.org/abs/1904.02480}
\BIBentrySTDinterwordspacing

\bibitem{Khoshelham2019SpatialMapping}
K.~{Khoshelham}, H.~{Tran}, and D.~{Acharya}, ``{Indoor Mapping Eyewear:
  Geometric Evaluation of Spatial Mapping Capability of Hololens},''
  \emph{ISPRS - International Archives of the Photogrammetry, Remote Sensing
  and Spatial Information Sciences}, vol. 4213, pp. 805--810, Jun. 2019.

\end{thebibliography}

\end{document}